# Does Normalization Methods Play a Role for Hyperspectral Image Classification?

Faxian Cao[1], Zhijing Yang[1*], Jinchang Ren[2], Mengying Jiang[1], Wing-Kuen Ling[1]

[1] School of Information Engineering, Guangdong University of Technology, Guangzhou, 510006, China

[2] Department of Electronic and Electrical Engineering, University of Strathclyde, Glasgow, G1 1XW, UK

*Email: yzhj@gdut.edu.cn

*Abstract*—For Hyperspectral image (HSI) datasets, each class have their salient feature and classifiers classify HSI datasets according to the class's saliency features, however, there will be different salient features when use different normalization method. In this letter, we report the effect on classifiers by different normalization methods and recommend the best normalization methods for classifier after analyzing the impact of different normalization methods on classifiers. Pavia University datasets, Indian Pines datasets and Kennedy Space Center datasets will apply to several typical classifiers in order to evaluate and analysis the impact of different normalization methods on typical classifiers.

I. Introduction

Hyperspectral image (HSI) classification is a challenging problem due to the fact that generally unfavorable ratio between the large number of spectral bands and the limited number of training samples, which results in the Hughes phenomenon [1]-[2]. In order to tackle this problem, a number of techniques have been proposed for feature extraction and dimensionality reduction [3], such as principal component analysis [4, 5] (PCA), singular spectrum analysis [6-9] (SSA) and segmented auto-encoder [10]. For data classification, typical approaches include support vector machine [11,12] (SVM), multi-kernel classification [13], *k*-nearest-neighbors [14] (*k*-NN) and multinomial logistic regression [15, 16] (MLR/LORSAL), Extreme Learning Machine [12, 17, 18, 19] (ELM) et al. These algorithms had achieved high classification accuracies and show good generalized performance, however, a pity that no one pay attention to the simplest part until now, which the data of HIS pre-processing.

As we know, the pixels of each class in HSI have salient feature, i.e., they have similar features if the pixels belonging to same class. So the data of pre-processing/normalization is a very important step for HSI classifiers. However, different methods of data pre-processing were be presents in different work and no one have analysed the impact of normalization on typical classifiers. In fact, different methods of normalization have certain impact on classifiers based on our investigation and evaluate. In this work, we analysis the impact of normalization on typical classifiers, we choose several typical classifiers, such as SVM [20], ELM [12] and LORSAL [15, 16] to evaluate the impact of normalization on these typical classifiers.

This paper is organized into four sections. In section II, we analysis the impact of different normalization method on HSI datasets/classifiers. Section III show the results of experiment. The conclusion are drawn in Section IV.

II. The impact of different normalization methods based on analysis

As mentioned before, different normalization methods have certain impact on classifiers, we will analysis the impact on classifiers in this section. In general, there are several normalization method:

(1) The first preprocessing method can be formulated as follows:

$$x_{ij}^* = x_{ij}/\max(x_{ij}) \qquad (1)$$

and we call it *Max*.

(2) The second method of normalization [21] can be view as

$$x_{ij}^* = (x_{ij} - \min(x_{ij}))/(\max(x_{ij}) - \min(x_{ij})) \qquad (2)$$

we call it *Max-Min*.

(3) The third one [22] is formulated as:
$$x_{ij}^* = x_{ij}/\text{bandmax}(x_{\cdot j}) \qquad (3)$$
we call it *Bandmax*.

(4) This method is similar to the one above
$$x_{ij}^* = (x_{ij} - \text{bandmin}(x_{\cdot j}))/(\text{bandmax}(x_{\cdot j}) - \text{bandmin}(x_{\cdot j})) \qquad (4)$$
we call it *Bandmax-min*.

(5) Let μ be the mathematical expectation of HSI data, σ be the standard deviation of HSI data, then the next method of normalization [2] can be formulated as:
$$x_i^* = (x_i \sim (\mu = 0, \sigma = 1)) \qquad (5)$$
we call it *Gaussian*.

(6) This method of normalization [23] can be given by following expression:
$$x_{ij}^* = x_{ij}/\sqrt{\sum \|x_i\|_2^2} \qquad (6)$$
we call it *Norm2*.

Have these in mind, due to the space, we choose widely used benchmarks HSI datasets the class Alfalfa of Indian Pines to analysis the impact of different normalization methods on classifiers. From Fig 1, we can see that there will be different shape of pixels in Alfalfa for different normalization methods. We also know that classifiers classify the HSI datasets according to the class's saliency features. There is no doubt that the performance of the classifier is affected by salient features of class when a class is transformed into different saliency features. So in next section, we will report the impact of different normalization on several typical classifiers.

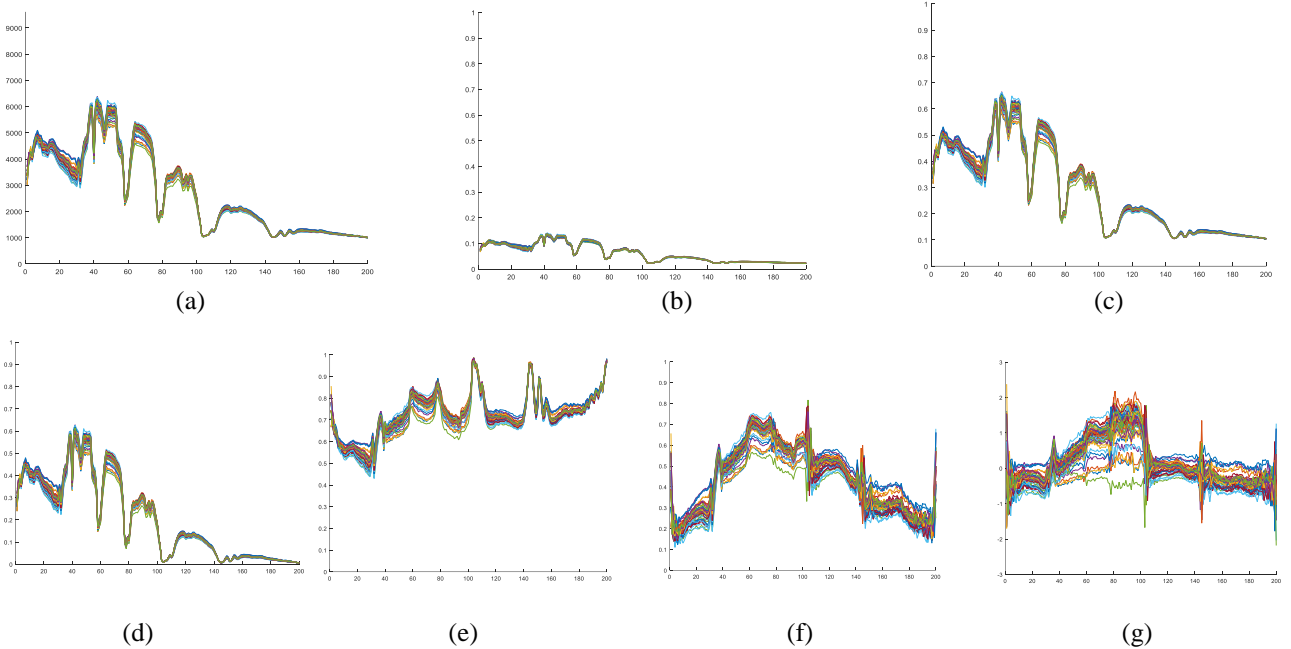

Fig 1. The impact of different normalization on Alfalfa of Indian Pines. (a) original spectral feature; (b) Norm2; (c) Max; (d) Max-min; (e) Bandmax; (f) Bandmax-min (g) Gaussian

III. Experiment Results

In this section, we evaluate the impact of different on different classifiers, such as SVM, ELM and LORSAL. And we use three HSI datasets to investigate the impact, the parameter of classifier mentioned before and two HSI datasets described as follows.

A. HSI datasets and parameter setting

(1) Pavia University: This dataset was obtained by the ROSIS instrument over the city of Pavia, Italy in 2001. This image scene corresponds to the true ground features of the University of Pavia, the dataset consists of $610 \times 340$ pixels and 115 spectral bands, and of which 103 bands are retained after discarding noisy and water absorption bands. And the spatial resolution is 1.3 m per pixel. The dataset includes 9 ground-truth classes and 42 776 labeled samples totally.

(2) Indian Pines: The dataset was obtained by the AVIRIS sensor in 1992. The image scene consists of 145 × 145 pixels and 220 spectral bands, of which 20 channels were discarded due to the atmospheric affection. The spatial resolution of this dataset is 20 m per pixel. And there are 16 classes and 10249 labeled samples in total in this dataset.

(3) Kennedy Space Center: The dataset was collected by NASA AVIRIS instrument over the Kennedy Space Center (KSC), Florida, in 1996. The KSC dataset, acquired from an altitude of approximately 20 km, have a spatial resolution of 18 m per pixel. which consists of 512×614 pixels and 224 bands of 10 nm width with center wavelengths from 400 - 2500 nm. After removing water absorption and low SNR bands, 176 bands were used for the analysis. And there are 13 classes and 5211 labeled samples totally.

Radial basis function (RBF) kernel was used for ELM and SVM and all the parameters of RBF all were performed by cross validation, the parameters of SVM and ELM had been describe in Zhou's work [12]. For LORSAL algorithm, the parameters setting had already been chosen in Sun's work [21], the parameters of LORSAL was setting to be 0.5, 0.85 and 0.35 for Kennedy Space Center datasets, Indian Pines datasets and Pavia University, respectively. In all of our experiments, the training samples are randomly selected in labeled samples and all labeled samples are used for testing. Tales 1 show the training samples and testing sample. In order to avoid the impact of parameters of algorithms and the random training samples on classification results, all of the classification results of these typical algorithms are all averaged by 100 Monte Carlo Runs.

Table 1. The training samples and testing samples for
Indian Pines, Kennedy Space Center and Pavia University

| Indian Pines | | | Kennedy Space Center | | | Pavia University | | |
|---|---|---|---|---|---|---|---|---|
| Class | Train | Test | Class | Train | Test | Class | Train | Test |
| Alfalfa | 6 | 54 | Scrub | 114 | 761 | Asphalt | 548 | 6631 |
| Corn-no till | 144 | 1434 | Willow swamp | 36 | 243 | Meadows | 548 | 18649 |
| Corn-min till | 84 | 834 | Cabbage palm hammock | 38 | 256 | Gravel | 392 | 2099 |
| Corn | 24 | 234 | Cabbage palm/oak hammock | 38 | 252 | Trees | 524 | 3064 |
| Grass/pasture | 50 | 497 | Slash pine | 24 | 161 | Metal sheets | 265 | 1345 |
| Grass/tree | 75 | 747 | Oak/broadleaf hammock | 34 | 229 | Bare soil | 532 | 5029 |
| Grass/pasture-mowed | 3 | 26 | Hardwood swamp | 16 | 105 | Bitumen | 375 | 1330 |
| Hay-windrowed | 49 | 489 | Graminoid marsh | 65 | 431 | Bricks | 514 | 3682 |
| Oats | 2 | 20 | Spartina marsh | 78 | 520 | Shadows | 231 | 947 |
| Soybeans-no till | 97 | 968 | Cattail marsh | 61 | 404 | | | |
| Soybeans-min till | 247 | 2468 | Salt marsh | 63 | 419 | | | |
| Soybeans-clean till | 62 | 614 | Mud flats | 75 | 503 | | | |
| Wheat | 22 | 212 | Water | 139 | 927 | | | |
| Woods | 130 | 1294 | | | | | | |
| Bldg-grass-tree-drives | 38 | 380 | | | | | | |
| Stone-steel towers | 10 | 95 | | | | | | |

B. The classification results of different typical classifiers

(1) Pavia University datasets:

From Table 2, 3 and 4. We can see that the norm2 normalization method have achieved worst classification accuracy compared with other normalization method . Max, Max-min, Bandmax and Bandmax-min have similar classification accuracy and Gaussian have lower classification accuracy than Max, Max-min, Bandmax and Bandmax-min.

(2) Indian Pines datasets:

We can see that Bandmax have achieved best classification accuracy for ELM, SVM and LORSAL in Table 1, 2 and 3. The classification results of bandmax-min are lower than bandmax. The classification accuracy become complex for other normalization method due to the complicate structure of Indian Pines and the different principle of other normalization methods.

(3) Kennedy Space Center:

From Table 2, 3 and 4, we can see that Gaussian method have achieved best results for ELM and LORSAL. Bandmax-min have achieved best classification for SVM. Bandmax and Bandmax-min have similar classification for ELM, SVM and LORSAL algorithms.

Based on the classification of Table 2, 3 and 4, we can conclude Bandmax are most good and stable normalization method compared with other normalization methods. This is due to its principle accord with the law of image. In each band, each data is divided by the maximum value of this band. However, in most of work [15, 16, 23, etc] of LORSAL, they all use norm2 method for HSI classification. The main reason that LORSAL can't achieved good classification results due to the bad performances of norn2 normalization method. The similar situation was happened in [12, 23, etc], they use norm2 method for SVM. In Zhou' work [12], they use norm2 method for ELM. Therefore, we recommend the Bandmax-min method for data preprocessing in future hyperspectral classification algorithm.

Table 2. Classification Accuracy of ELM

The best results are shown in bold

| Data sets | | Norm2 | Max | Max-Min | Bandmax | Bandmax-min | Gaussian |
|---|---|---|---|---|---|---|---|
| Pavia University | OA | 0.8982 | 0.9351 | 0.9347 | 0.9348 | **0.9352** | 0.9298 |
| | AA | 0.9159 | **0.9446** | 0.9443 | 0.9443 | **0.9446** | 0.9389 |
| | kappa | 0.8672 | 0.9149 | 0.9144 | 0.9145 | **0.9151** | 0.9080 |
| Indian Pines | OA | 0.8714 | 0.8674 | 0.8673 | **0.8825** | 0.8570 | 0.8538 |
| | AA | 0.8244 | 0.8211 | 0.8228 | **0.8487** | 0.7951 | 0.7890 |
| | kappa | 0.8530 | 0.8485 | 0.8483 | **0.8658** | 0.8365 | 0.8327 |
| Kennedy Space Center | OA | 0.9262 | 0.9154 | 0.9154 | 0.9303 | 0.9309 | **0.9365** |
| | AA | 0.9178 | 0.8665 | 0.8665 | 0.8923 | 0.8931 | **0.9019** |
| | kappa | 0.8906 | 0.9056 | 0.9056 | 0.9224 | 0.9230 | **0.9292** |

Table 3. Classification Accuracy of SVM

The best results are shown in bold

| Data sets | | Norm2 | Max | Max-Min | Bandmax | Bandmax-min | Gaussian |
|---|---|---|---|---|---|---|---|
| Pavia University | OA | 0.9006 | **0.9395** | **0.9395** | **0.9395** | **0.9395** | 0.9355 |
| | AA | 0.9194 | **0.9472** | **0.9472** | **0.9472** | **0.9472** | 0.9417 |
| | kappa | 0.8704 | **0.9205** | **0.9205** | **0.9205** | **0.9205** | 0.9154 |
| Indian Pines | OA | 0.8736 | 0.8672 | 0.867 | **0.8833** | 0.8481 | 0.8419 |
| | AA | 0.8536 | 0.8414 | 0.8407 | **0.858** | 0.8039 | 0.7932 |
| | kappa | 0.8559 | 0.8485 | 0.8483 | **0.8669** | 0.8265 | 0.8193 |
| Kennedy Space Center | OA | 0.9346 | 0.9363 | 0.9363 | 0.9411 | **0.9412** | 0.9366 |
| | AA | 0.9062 | 0.9045 | 0.9045 | **0.9085** | 0.9084 | 0.8992 |
| | kappa | 0.9271 | 0.9291 | 0.9291 | 0.9344 | **0.9345** | 0.9293 |

Table 4. Classification Accuracy of LORSAL

| Data sets | | Norm2 | Max | Max-Min | Bandmax | Bandmax-min | Gaussian |
|---|---|---|---|---|---|---|---|
| Pavia University | OA | 0.8895 | 0.9334 | 0.9334 | 0.9334 | **0.9335** | 0.9305 |
| | AA | 0.8991 | 0.9402 | 0.9402 | 0.9402 | **0.9403** | 0.9352 |
| | kappa | 0.8559 | 0.9127 | 0.9127 | 0.9127 | **0.9128** | 0.9088 |
| Indian Pines | OA | 0.8426 | 0.8495 | 0.8495 | **0.8638** | 0.8368 | 0.8329 |
| | AA | 0.7847 | 0.7849 | 0.7849 | **0.8099** | 0.7521 | 0.8091 |
| | kappa | 0.8203 | 0.8281 | 0.8281 | **0.8445** | 0.8136 | 0.7500 |
| Kennedy | OA | 0.9250 | 0.8460 | 0.8460 | 0.9309 | 0.9310 | **0.9339** |

| Space  | AA    | 0.8901 | 0.7524 | 0.7524 | 0.8966 | 0.8967 | **0.9006** |
| Center | kappa | 0.9164 | 0.8279 | 0.8279 | 0.9231 | 0.9231 | **0.9264** |

## IV. Conclusions

In this work, we find an interest phenomenon that these are the impact of normalization methods on classification results. We display the shape of pixels after using different normalization and find the impact on classification results based on lots of experiment. We recommend the Band-Maxmin method for Hyperspectral image classification.